\documentclass{article}
\usepackage[preprint]{log_2022}

\usepackage{booktabs}            
\usepackage{multirow}            
\usepackage{amsfonts}            
\usepackage{graphicx}            
\usepackage{duckuments}          

\usepackage[numbers,compress,sort]{natbib}

\usepackage{adjustbox}
\usepackage{makecell}
\usepackage{multirow}
\usepackage[ruled,vlined,linesnumbered]{algorithm2e}
\usepackage{enumitem}
\usepackage{subcaption}
\usepackage{amsmath,amsthm}
\usepackage{xcolor}
\usepackage{diagbox}
\usepackage{mdframed}
\newmdtheoremenv[%
  backgroundcolor=gray!20,
  linecolor=red!60!black,
  linewidth=2pt,
  topline=false,
  rightline=false,
  skipabove=10pt,
  skipbelow=10pt,
  leftline=false]{regbox}{Box}

\newcommand{\eg}{\emph{e.g.}\xspace}

\newcommand{\GCN}{\texttt{GCN}\xspace}
\newcommand{\GIN}{\texttt{GIN}\xspace}
\newcommand{\GNNExplainer}{\texttt{GNNExplainer}\xspace}
\newcommand{\PGExplainer}{\texttt{PGExplainer}\xspace}
\newcommand{\MUTAG}{\texttt{MUTAG}\xspace}
\newcommand{\BAmotifs}{\texttt{BA-2motifs}\xspace}
\newcommand{\BAMultiShapes}{\texttt{BA-Multi-Shapes}\xspace}
\newcommand{\GraphTwitter}{\texttt{Graph-Twitter}\xspace}
\newcommand{\ACC}{\ensuremath{\mathsf{Acc}}\xspace}
\newcommand{\FIDP}{\ensuremath{\mathsf{Fid}^{+}}\xspace}
\newcommand{\FIDN}{\ensuremath{\mathsf{Fid}^{-}}\xspace}
\title[On the Robustness of Post-hoc GNN Explainers to Label Noise]
{On the Robustness of Post-hoc GNN Explainers to Label Noise}

\author[Z. Zhong et al.]
{%
Zhiqiang Zhong\\
\institute{Aarhus University}\\
\email{zzhong@cs.au.dk}
\And
Yangqianzi Jiang\\
\institute{Rice University}\\
\email{yj50@rice.edu}
\And
Davide Mottin\\
\institute{Aarhus University}\\
\email{davide@cs.au.dk}
}

\begin{document}

\maketitle

\begin{abstract} 
Proposed as a solution to the inherent black-box limitations of graph neural networks (GNNs), \emph{post-hoc GNN explainers} aim to provide precise and insightful explanations of the behaviours exhibited by trained GNNs. 
Despite their recent notable advancements in academic and industrial contexts, the \emph{robustness} of post-hoc GNN explainers remains unexplored when confronted with label noise. 
To bridge this gap, we conduct a systematic \emph{empirical investigation} to evaluate the efficacy of diverse post-hoc GNN explainers under varying degrees of label noise.
Our results reveal several key insights: Firstly, post-hoc GNN explainers are susceptible to label perturbations. 
Secondly, even minor levels of label noise, inconsequential to GNN performance, harm the quality of generated explanations substantially.
Lastly, we engage in a discourse regarding the progressive recovery of explanation effectiveness with escalating noise levels.

\end{abstract}

\section{Introduction} 
\label{sec:introduction}
The emergence of Graph Neural Networks (GNNs) has revolutionised machine learning on graph-structured data~\cite{WPCLZY21,J22,ZLP23}.
Nevertheless, a substantial concern has been raised within the community: GNN models can be easily manipulated/attacked~\cite{ZAG18,DLTHWZS18} by unnoticeable modifications. 
To counter this, researchers proposed robust GNN models against diverse adversarial attacks~\cite{ZG19,BG19}. 
However, a significant gap persists as current GNNs struggle to provide insightful interpretations of their underlying mechanisms and outputs.
To tackle this limitation, recent researchers proposed post-hoc GNN explainers, designed to explain the behaviour of a trained GNN models~\cite{YYGJ23,YBYZL19,LCXYZCZ20,ST21,PKRMH19,SELNSMM22}. 

While the robustness of GNNs is a well studied phenomenon, that of post-hoc explainers has been overlooked. 
As such, \emph{we question how robust are post-hoc GNN explainers in the face of label noise.}
To this end, we pose two related research questions: \emph{(i) Are post-hoc GNN explainers robust to malicious label attacks?} and \emph{(ii) Does the robustness of GNN models unequivocally guarantee the effectiveness of post-hoc explainers?}

In pursuit of answers to these questions, we conduct an empirical investigation. 
Our focus centers on investigating the impact of a widely existing noise form, namely label noise, on post-hoc GNN explainers within the context of graph classification. 
We integrate two benchmark post-hoc explainers (\textsc{GNNExplainer~\cite{YBYZL19}} and \textsc{PGExplainer~\cite{LCXYZCZ20}}) into a unified evaluation framework and carefully evaluate the effectiveness of explanations across four graph datasets, including two real-world datasets of different topics and two synthetic datasets.

The outcomes of our study effectively answered the raised questions. 
Firstly, the selected benchmark GNN explainers prove lacking in robustness against label noise, evidenced by the substantial decline in explanation quality upon random graph label disturbances. 
%
Second, we observe that the effectiveness of GNN explainers is severely compromised, even with minor levels of label noise, despite the robust performance retained by the GNN models.
%
Besides, we discuss the impracticality of one current metric for evaluating explanations within the context of post-hoc explainer robustness analysis since it arrives at optimal values while feeding with ambiguous labels. 
An additional noteworthy: beyond a noise threshold of 50\%, explanation effectiveness gradually recovers to levels comparable to those without noise as noise levels continue escalating.
We illustrate this with specific explanation instances, showcasing that inverted label signals enable GNN explainers to discern important features.


\section{Preliminaries} 
\label{sec:preliminaries}
\textbf{GNNs and GNN Explainers.}
Graph neural networks (GNNs)~\cite{WPCLZY21,J22} have emerged as a powerful class of deep learning models designed to handle data structured as graph, making them invaluable in various domains, \eg, social network~\cite{CCZP222}, scientific discovery~\cite{WFDG23} and biology~\cite{ZBM23}.
Given a graph $\mathcal{G} = (\mathcal{V}, \mathcal{E})$ with $n$ nodes and corresponding node attributes $\mathbf{X} \in \mathbb{R}^{n \times d}$. 
GNNs can learn to generate effective property prediction across nodes, edges, and graphs.
For instance, graph $\mathcal{G}$'s label is predicted as $\hat{y} = \arg\max_y (\mathrm{GNN}_{\theta} (\mathbf{A}, \mathbf{X}))$, where $\mathbf{A} \in \{0, 1\}^{n \times n}$ is the adjacency matrix summarising $\mathcal{V}$ and $\mathcal{E}$ and $\theta$ is the set of trainable parameters of GNN model. 

In response to the black-box limitations of GNNs, a range of GNN explainers have been introduced~\cite{YBYZL19,LCXYZCZ20,DW21,ZLWLL22}. 
Broadly categorised as \emph{self-explainable} and \emph{post-hoc} models~\cite{YYGJ23}, these GNN explainers produce interpretations either during or after GNN model training. 
Within this context, our study focuses on post-hoc GNN explainers, which generate explanations $\mathbf{E}$ based on trained GNN, generated explanations and graph.

\textbf{Evaluation of Post-hoc GNN Explainers.}
We utilise two popular label-agnostic evaluation metrics: \emph{fidelity+} (\FIDP) and \emph{fidelity-} (\FIDN)~\cite{YYGJ23}. 
\emph{fidelity+} measures the prediction change following the removal of relevant features; \emph{fidelity-} assesses the change by retaining only the relevant features:
\begin{equation}
\label{eq:fidelity}
\FIDP{=}\frac{1}{N} \sum_{i=1}^{N} (\mathrm{GNN}_{\theta}(\mathcal{G}_i)_{\hat{y}_i} {-} \mathrm{GNN}_{\theta}(\mathcal{G}_i^{1-\mathbf{E}_i})_{\hat{y}_i});
\;\;
\FIDN{=}\frac{1}{N} \sum_{i=1}^{N} (\mathrm{GNN}_{\theta}(\mathcal{G}_i)_{\hat{y}_i} {-} \mathrm{GNN}_{\theta}(\mathcal{G}_i^{\mathbf{E}_i})_{\hat{y}_i})
\end{equation}
where $\hat{y}_i$ is the predicted prediction of graph $\mathcal{G}_i$ and $\mathcal{G}_i^{\mathbf{E}_i}$ represents the new graph obtained by keeping features of $\mathcal{G}_i$ based on the mask $\mathbf{E}_i$

\textbf{Robustness of Post-hoc GNN Explainers.}
While the remarkable efficacy of GNNs has attracted considerable academic attention, their fragile performance on the maliciously manipulated graph also caused researchers' concerns~\cite{ZAG18,DLTHWZS18}. 
Consequently, increasing the robustness of GNNs on face to adversarial attacks has also been studied~\cite{ZG19,BG19}. 
However, the post-hoc GNN explainer's robustness has largely remained unexplored within the research community. 
We hereby pose two critical research questions: \textbf{Q1:} \emph{Can post-hoc GNN explainers withstand malicious attacks?} and \textbf{Q2:} \emph{Does the robustness of GNN models unequivocally extend to stable \textit{fidelity} of post-hoc explainers?}


\section{Empirical Study} 
\label{sec:empirical_study}
To address the aforementioned research questions, we evaluate the explanation quality in terms of \FIDP and \FIDN of two benchmark post-hoc GNN explainers, \GNNExplainer~\cite{YBYZL19} and \PGExplainer~\cite{LCXYZCZ20}, on two GNN models, \GCN~\cite{KW17} and \GIN~\cite{XHLJ19}. 
This study spans across four graph datasets, two real-world datasets, \MUTAG and \GraphTwitter, and two synthetic, \BAmotifs and \BAMultiShapes.
A detailed overview of these datasets, along with statistical information can be found in Table~\ref{table:summary_dataset} in the Appendix~\ref{sec:appendix}.

\textbf{Implementation.}
We integrate the implementations of \GCN and \GIN from PyG~\cite{FL19} and \GNNExplainer and \PGExplainer from the original papers into a unified framework built with DIG~\cite{LLWX21}. 
Graph classification performance is evaluated by classification accuracy (\ACC), and generated explanations are measured by \FIDP and \FIDN.
We select the best hyper-parameters of GNN model and explainers follow the benchmark settings of DIG. 

\textbf{Noisy Label Generation.} We first select a set of training graph indexes according to the noisy level ($\lambda$). 
Then, if the graph original label is $0$, we randomly select a value from $\{1, 2 \}$ to replace its label. 
We do not change the label of test graphs. 

\subsection{Results}
\label{subsec:results}

\begin{figure}[!ht]
\centering
\includegraphics[width=1.\linewidth]{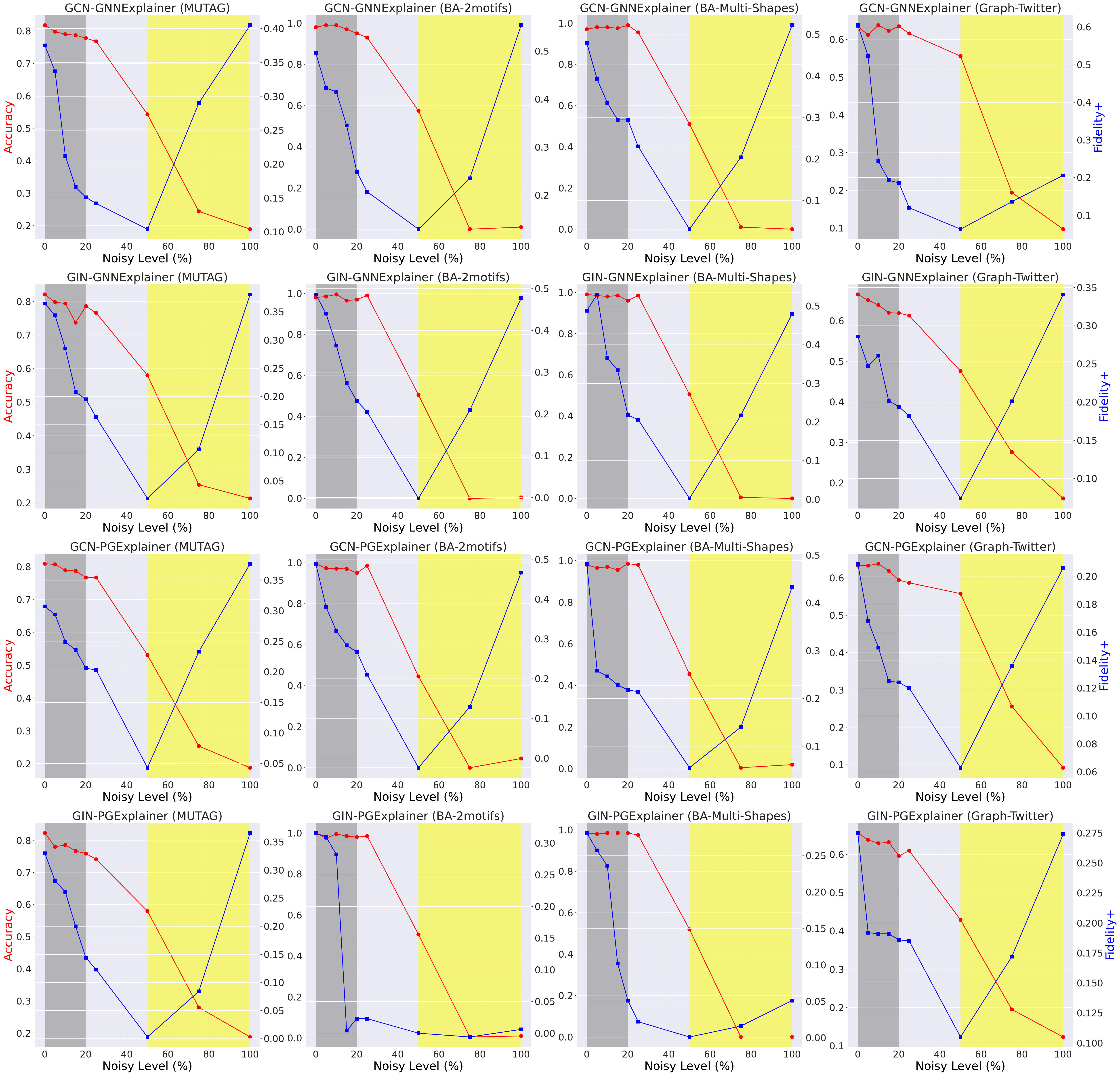}
\caption{
The performance of GNN models and explainers of different label noisy levels. 
}
\label{fig:performance}
\vspace{-5mm}
\end{figure}

Due to space limit, the detailed evaluation (average results of three runs) is displayed in Table~\ref{table:macro_evaluation_gnn_explainers}-\ref{table:micro_evaluation_gnn_explainers} of Appendix~\ref{sec:appendix}. 
To facilitate the readers to understand our results, we partially summarise them into Figure~\ref{fig:performance}. 
\textcolor{gray}{\textbf{Dark grey}} and \textcolor{Goldenrod}{\textbf{yellow}} shadows highlight the regions of $\lambda \leq 20\%$ and $\lambda \geq 50 \%$, respectively. 

\textbf{Q1.} \textbf{Post-hoc GNN explainers are susceptible to label noise.}
Figure~\ref{fig:performance} emphasises that both \ACC and \FIDP are significantly impacted by varying levels of label noise. 
The fluctuating trends of \FIDP (blue lines) underscore the instability of explanation quality, whereas the trends of \ACC (red lines) echo the findings about GNN robustness discussed in previous work~\cite{ZAG18,DLTHWZS18}.

\textbf{Observation 1.} 
From the results in Table~\ref{table:macro_evaluation_gnn_explainers}-\ref{table:micro_evaluation_gnn_explainers}, we find out that \FIDN decreases as $\lambda$ increases from $0\%$ to $50\%$. 
However, the definition of \FIDN indicating lower values as more satisfactory contradicts this outcome. 
In our scenario, confusing labels lead to ambiguous predictions, subsequently causing $\FIDN \to 0$. 
We thereby argue that \emph{\FIDN proves unsuitable as a valid metric within the context of investigating post-hoc GNN explainer robustness.}

\textbf{Q2.} \textbf{The robustness of GNN models does not extend to the stable fidelity of post-hoc explainers.}
Within the grey shadow regions of Figure~\ref{fig:performance}, it is evident that \ACC remains relatively stable, indicating that GNN models exhibit robustness in the face of minor noisy levels. 
Conversely, \FIDP experiences substantial drops at the same noise levels ($\lambda \leq 20\%$), revealing the heightened sensitivity of different post-hoc explainers to even minor noisy levels of label noise.

\begin{figure}[!ht]
\centering
\includegraphics[width=1.\linewidth]{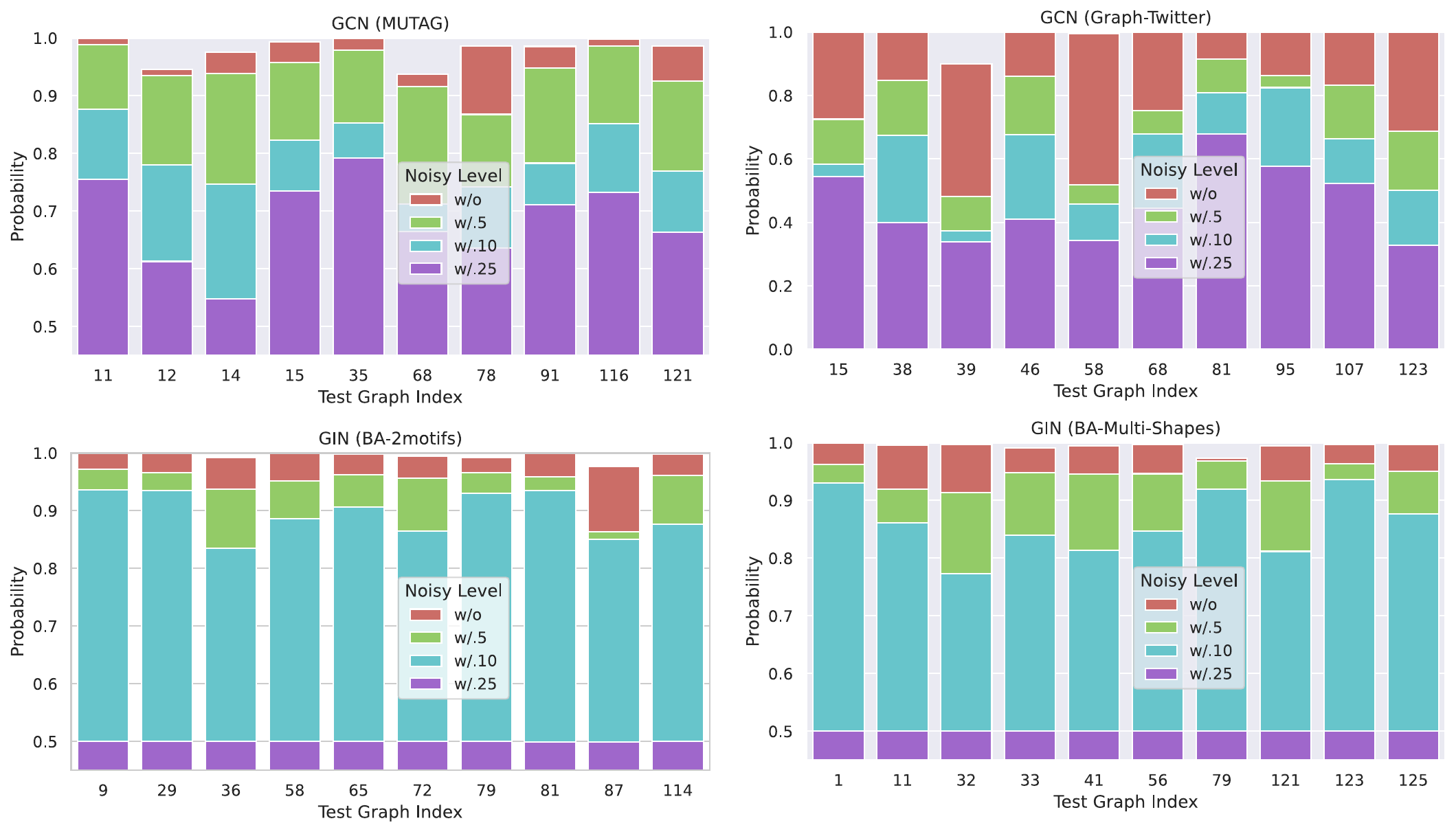}
\caption{
Example predicted probabilities of GNNs on four datasets of different label noisy levels. 
}
\label{fig:prediction_confidence}
\vspace{-5mm}
\end{figure}

\textbf{Observation 2.}
To unveil the grey shadow regions, we present the corresponding predicted probabilities of the true label of ten example test graphs in Figure~\ref{fig:prediction_confidence}. 
Although GNN models manage to accurately classify these graphs with minor noisy levels, yet, the predicted probabilities are affected by introduced noises, which are not represented in \ACC. 
In contrast, these predicted probabilities would be passed to GNN explainers as essential inputs to generate explanations, as illustrated in Figure~\ref{fig:post-hoc_gnnexplainer_mechanism}. 
We posit that this might the chief reason for the contrasting performance of \ACC and \FIDP. 

\begin{figure}[!ht]
\centering
\includegraphics[width=1.\linewidth]{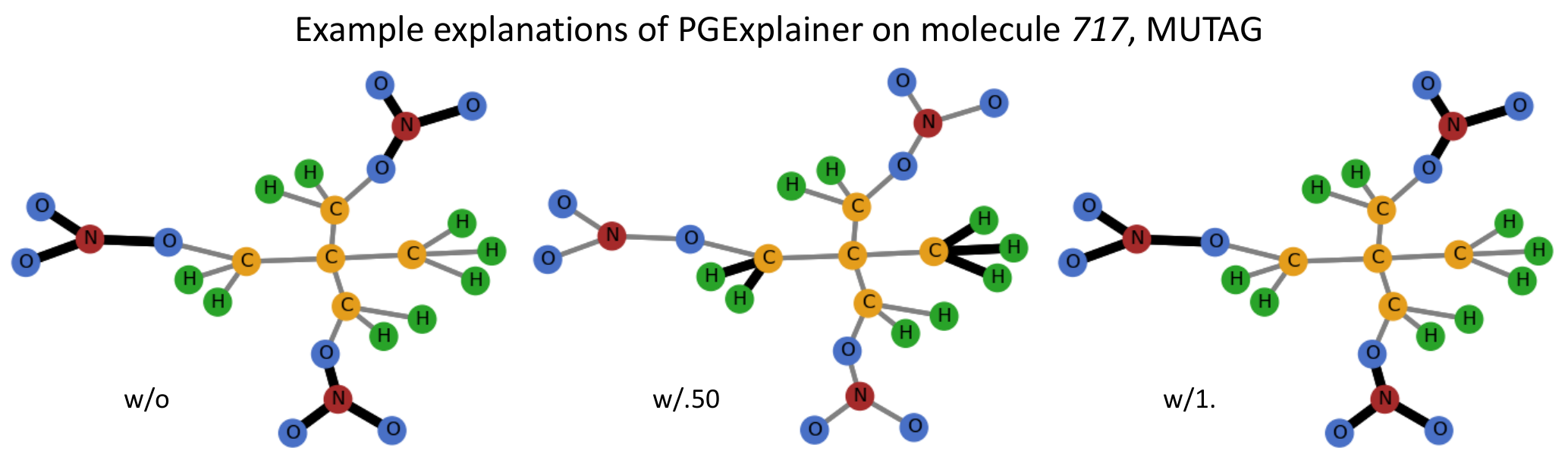}
\caption{
Explanations (bold edges) generated by \PGExplainer for molecule graph 717 of \MUTAG. 
}
\label{fig:example_explanations}
\vspace{-5mm}
\end{figure}

\textbf{Observation 3.}
Another interesting phenomena we observed in Figure~\ref{fig:performance} is that \emph{beyond a noise threshold of $\lambda=$50\%, \FIDP gradually returns to levels comparable to those without noise as noise levels continue to escalate.}
To better understand this phenomenon, we present the generated explanations about an example molecule graph (id = 717) in Figure~\ref{fig:example_explanations}. 
At $\lambda=$50\%, \PGExplainer fails to identity key motifs (NO2), yet sucessfully does so at $\lambda=$100\%. 
This suggests that \emph{confusing label signals mislead GNN models and explainers, while inverted label signals enables GNN models to predict reverse labels while identifying important features}.


\section{Concluding Remarks and Future Directions}
\label{sec:conclusion}
This extended abstract represents, to the best of our knowledge, the first preliminary exploration into the robustness of post-hoc GNN explainers against label noise. 
Our findings introduce several interesting research questions to the community: Firstly, we establish the susceptibility of post-hoc GNN explainers to label noise. 
Secondly, our investigation highlights that the \textit{fidelity} of post-hoc explainers can be significantly impacted by minor noise, which does not conduce a noticeable influence on the GNN model's performance. 
This underscores the complexity inherent in bolstering the robustness of post-hoc GNN explainers, necessitating dedicated efforts. 
Additionally, our study unveils the impracticability of \FIDN metric for explainer robustness study since it naturally gets optimised with high noisy levels. 
In follow-up work, there are several promising future directions to explore.
For instance, designing robust post-hoc GNN explainers to label noise, refining explanation evaluation metrics for comprehensive measurement, and developing large-scale benchmark datasets. 

\section*{Acknowledgements}
This work is supported by the Horizon Europe and Innovation Fund Denmark under the Eureka, Eurostar grant no E115712 - AAVanguard.

\bibliographystyle{unsrtnat}
\bibliography{full_format_references}

\newpage
\appendix
\section{Supplement Material}
\label{sec:appendix}

\begin{table}[!ht]
\caption{
    Summary of datasets used in our empirical study.
}
\label{table:summary_dataset}
\setlength{\tabcolsep}{3pt} 
\centering
\footnotesize
\begin{tabular}{l|r|r|r|r|r|r}
\toprule 
\textsc{Name} & \textsc{Cat.} & \textsc{\#Graphs} & \textsc{\#Nodes (Avg.)} & \textsc{\#Edges (avg)} & \textsc{\# Node Feat.} & \textsc{\#Classes} \\
\midrule
\MUTAG & Real & 4,337  & 30.3 & 61.5 & 14 & 2 \\
\BAmotifs & Syn. & 1,000  & 25 & 51 & 10 & 2 \\
\BAMultiShapes & Syn. & 1,000  & 25 & 51 & 10 & 2 \\
\GraphTwitter & Real & 6,940 & 21.1 & 40.2 & 768 & 3  \\ 
\bottomrule
\end{tabular}
\end{table}

\textbf{Datasets.}
We consider four graph classification datasets as summarised in Table~\ref{table:summary_dataset}. 
\MUTAG is a real-world dataset of 4,337 molecule graphs labelled according to their mutagenic effect~\cite{YBYZL19}. 
\GraphTwitter is a real-world sentiment graph dataset of 6,940 text graphs, which is constructed based on text sentiment analysis~\cite{YYGJ23}. 
\BAmotifs~\cite{LCXYZCZ20} and \BAMultiShapes~\cite{ALBLP22} are two synthetic datasets of 1,000 random Barabasi-Albert (BA) graphs. 
Each graph of \BAmotifs is obtained by attaching either HouseMotif or CycleMotif. 
Each graph of \BAMultiShapes is obtained by attaching one of HouseMotif, WheelMotif and GridMotif. 
These graphs are assigned to one of the two classes according to the type of attached motifs.
We split datasets into train/valid/test ($70\%$, $10\%$, $20\%$) subsets for the experiments. 
GNN models are trained on train datasets and test on test datasets. 

\textbf{Implementation Details.}
Our implementations mainly follows the settings of officially public Pytorch code of \PGExplainer~\cite{LCXYZCZ20} (\url{https://github.com/divelab/DIG/tree/main/dig/xgraph/PGExplainer}). 
Particularly, we first train a GNN model (two-layers or three-layers of fixed hidden dimension 128) and select the one with the best \ACC. 
After, we pass trained GNN model, obtained predictions and the graph to the GNN explainers to ontain explanations and compute their \FIDP and \FIDN following \url{https://github.com/divelab/DIG/tree/dig/benchmarks/xgraph}. 
The learning rate to train GNN models and explainers is fixed as $0.001$. 
Other settings follow the default their official implementations. 

\begin{table}[!ht]
\caption{
    Evaluation of GNN models, \textsc{GNNExplainer} and \textsc{PGExplainer} on real-world datasets, with (w/) and without (w/o) label noise.
    For instance, $\lambda = $w/0.25 indicates 25\% training node labels are randomly manipulated. 
    Model performance is evaluated on classification accuracy (\ACC) and explanation quality is evaluated on fidelity+ (\FIDP) and fidelity- (\FIDN) on test dataset. 
}
\label{table:macro_evaluation_gnn_explainers}
\setlength{\tabcolsep}{3pt} 
\centering
\resizebox{1.\linewidth}{!}{
\begin{tabular}{l|c|cc|cc|cc|cc|cc|cc}
\toprule 
\textbf{Dataset} & $\lambda$ & \multicolumn{12}{c}{\textbf{\textsc{GNNExplainer}}} \\
\midrule
& & \multicolumn{6}{c|}{\GCN} & \multicolumn{6}{c}{\GIN} \\
& & \multicolumn{2}{c|}{\ACC} & \multicolumn{2}{c|}{\FIDP} & \multicolumn{2}{c|}{\FIDN} & \multicolumn{2}{c|}{\ACC} & \multicolumn{2}{c|}{\FIDP} & \multicolumn{2}{c|}{\FIDN} \\
\midrule
\midrule
\multirow{5}{*}{\MUTAG} 
& w/o 
& 0.818 & - & 0.375 & - & 0.425 & - 
& 0.821 & - & 0.365 & - & 0.367 & - \\
& w/.25 
& 0.768 & -6.1\% & 0.142 & -62.1\% & 0.144 & -66.1\%
& 0.766 & -6.7\% & 0.163 & -55.3\% & 0.161 & -56.1\% \\
& w/.50 
& 0.543 & -33.6\% & 0.104 & -72.3\% & 0.098 & -76.9\%
& 0.580 & -29.4\% & 0.019 & -94.8\% & 0.019 & -94.8\%\\
& w/.75 
& 0.244 & -70.2\% & 0.290 & -22.7\% & 0.288 & -32.2\%
& 0.254 & -69.1\% & 0.106 & -71.0\% & 0.110 & -70.0\% \\
& w/1. 
& 0.189 & -76.9\% & 0.405 & 8.0\% & 0.399 & -6.1\%
& 0.213 & -74.1\% & 0.381 & 4.4\% & 0.381 & 3.8\% \\


\midrule
\multirow{5}{*}{\BAmotifs} 
& w/o 
& 0.980 & - & 0.496 & - & 0.496 & - 
& 0.980 & - & 0.486 & - & 0.486 & - \\
& w/.25 
& 0.930 & -5.1\% & 0.207 & -58.3\% & 0.221 & -55.4\%
& 0.990 & 1.0\% & 0.205 & -57.8\% & 0.204 & -58.0\% \\
& w/.50 
& 0.575 & -41.3\% & 0.129 & -74.0\% & 0.164 & -66.9\%
& 0.505 & -48.5\% & -0.002 & -100.4\% & -0.002 & -100.4\% \\
& w/.75 
& 0.000 & -100.0\% & 0.235 & -52.6\% & 0.235 & --52.6\%
& 0.000 & -100.0\% & 0.209 & -57.0\% & 0.200 & -58.8\% \\
& w/1. 
& 0.010 & -99.0\% & 0.554 & 11.7\% & 0.556 & 12.1\%
& 0.005 & -99.5\% & 0.477 & -1.9\% & 0.476 & -2.1\% \\

\midrule
\multirow{5}{*}{\texttt{BAMult.S.}} 
& w/o 
& 0.970 & - & 0.479 & - & 0.479 & - 
& 0.990 & - & 0.488 & - & 0.488 & - \\
& w/.25 
& 0.955 & -1.5\% & 0.230 & -52.0\% & 0.223 & -53.4\%
& 0.985 & -0.5\% & 0.206 & -57.8\% & 0.202 & -58.6\% \\
& w/.50 
& 0.510 & -47.4\% & 0.031 & -93.5\% & 0.050 & -89.6\%
& 0.505 & -49.0\% & 0.002 & -99.6\% & 0.002 & -99.6\% \\
& w/.75 
& 0.010 & -99.0\% & 0.204 & -57.4\% & 0.198 & -58.7\% 
& 0.005 & -99.5\% & 0.217 & -55.5\% & 0.212 & -56.6\%  \\
& w/1. 
& 0.000 & -100.0\% & 0.522 & 9.0\% & 0.528 & 10.2\%
& 0.000 & -100.0\% & 0.480 & -1.6\% & 0.479 & -1.8\%  \\

\midrule
\multirow{5}{*}{\texttt{G.-Twitter}} 
& w/o 
& 0.635 & - & 0.605 & - & 0.594 & - 
& 0.665 & - & 0.286 & - & 0.262 & - \\
& w/.25 
& 0.616 & -3.0\% & 0.161 & -73.4\% & 0.162 & -72.7\% 
& 0.613 & -7.8\% & 0.182 & -36.4\% & 0.173 & -34.0\% \\ 
& w/.50 
& 0.556 & -12.4\% & 0.055 & -90.9\% & 0.054 & -90.9\%
& 0.476 & -28.4\% & 0.074 & -74.1\% & 0.070 & -73.3\% \\
& w/.75 
& 0.194 & -69.4\% & 0.012 & -98.0\% & 0.012 & -98.0\%
& 0.276 & -58.5\% & 0.201 & -29.7\% & 0.197 & -24.8\% \\
& w/1. 
& 0.097 & -84.7\% & 0.121 & -80.0\% & 0.120 & -79.8\%
& 0.162 & -75.6\% & 0.341 & 19.2\% & 0.309 & 17.9\% \\

\midrule
\textbf{Dataset} & $\lambda$ & \multicolumn{12}{c}{\textbf{\textsc{PGExplainer}}} \\
\midrule
\midrule

\multirow{5}{*}{\MUTAG} 
& w/o 
& 0.809 & - & 0.307 & - & 0.431 & - 
& 0.823 & - & 0.330 & - & 0.482 & - \\
& w/.25 
& 0.767 & -5.2\% & 0.203 & -33.9\% & 0.274 & -36.4\%
& 0.741 & -10.0\% & 0.123 & -62.7\% & 0.117 & -75.7\% \\
& w/.50 
& 0.531 & -34.4\% & 0.043 & -86.0\% & 0.094 & -78.2\%
& 0.580 & -29.5\% & 0.003 & -99.1\% & 0.013 & -97.3\% \\
& w/.75 
& 0.254 & -68.6\% & 0.233 & -24.1\% & 0.287 & -33.4\%
& 0.280 & -66.0\% & 0.084 & -74.5\% & 0.041 & -91.5\% \\
& w/1. 
& 0.188 & -76.8\% & 0.377 & 22.8\% & 0.443 & 2.8\%
& 0.188 & -77.2\% & 0.366 & 10.9\% & 0.406 & -15.8\% \\


\midrule
\multirow{5}{*}{\BAmotifs} 
& w/o 
& 0.995 & - & 0.489 & - & 0.489 & - 
& 1.000 & - & 0.316 & - & 0.481 & - \\
& w/.25 
& 0.985 & -1.0\% & 0.210 & -57.1\% & 0.213 & -56.4\%
& 0.985 & -1.5\% & 0.024 & -92.4\% & 0.210 & -56.3\% \\
& w/.50 
& 0.445 & -55.3\% & -0.024 & -104.9\% & 0.193 & -60.5\%
& 0.505 & -49.5\% & -0.000 & -100.0\% & -0.000 & -100.0\% \\
& w/.75 
& 0.000 & -100.0\% & 0.129 & -73.6\% & 0.147 & -69.9\%
& 0.005 & -99.5\% & -0.006 & -101.9\% & 0.209 & -56.5\% \\
& w/1. 
& 0.045 & -95.5\% & 0.467 & -4.5\% & 0.489 & 0.0\%
& 0.010 & -99.0\% & 0.006 & -98.1\% & 0.463 & -3.7\% \\

\midrule
\multirow{5}{*}{\texttt{BAMult.S.}} 
& w/o 
& 0.980 & - & 0.482 & - & 0.482 & - 
& 0.985 & - & 0.281 & - & 0.485 & - \\
& w/.25 
& 0.980 & 0.0\% & 0.214 & -55.6\% & 0.252 & -47.7\%
& 0.975 & -1.0\% & 0.022 & -92.2\% & 0.143 & -70.5\% \\
& w/.50 
& 0.455 & -53.6\% & 0.055 & -88.6\% & 0.108 & -77.6\% 
& 0.520 & -47.2\% & 0.001 & -99.6\% & 0.002 & -99.6\% \\
& w/.75 
& 0.005 & -99.5\% & 0.140 & -71.0\% & 0.144 & -70.1\%
& 0.000 & -100.0\% & 0.016 & -94.3\% & 0.237 & -51.1\% \\
& w/1. 
& 0.020 & -98.0\% & 0.433 & -10.2\% & 0.443 & -8.1\%
& 0.000 & -100.0\% & 0.051 & -81.9\% & 0.501 & 3.3\% \\

\midrule
\multirow{5}{*}{\texttt{G.-Twitter}} 
& w/o 
& 0.633 & - & 0.209 & - & 0.333 & - 
& 0.656 & - & 0.275 & - & 0.255 & - \\
& w/.25 
& 0.587 & -7.3\% & 0.120 & -42.6\% & 0.157 & -52.9\% 
& 0.610 & -7.0\% & 0.185 & -32.7\% & 0.174 & -31.8\% \\
& w/.50 
& 0.558 & -11.8\% & 0.063 & -69.9\% & 0.062 & -81.4\%
& 0.429 & -34.6\% & 0.105 & -61.8\% & 0.030 & -88.2\% \\
& w/.75 
& 0.256 & -59.6\% & 0.136 & -34.9\% & 0.263 & -21.0\% 
& 0.195 & -70.3\% & 0.172 & -37.5\% & 0.172 & -32.5\% \\
& w/1. 
& 0.092 & -85.5\% & 0.206 & -1.4\% & 0.307 & -7.8\%
& 0.123 & -81.3\% & 0.274 & -0.4\% & 0.276 & 8.2\% \\

\bottomrule
\end{tabular}
}
\end{table}

\begin{figure}[!ht]
\centering
\includegraphics[width=1.\linewidth]{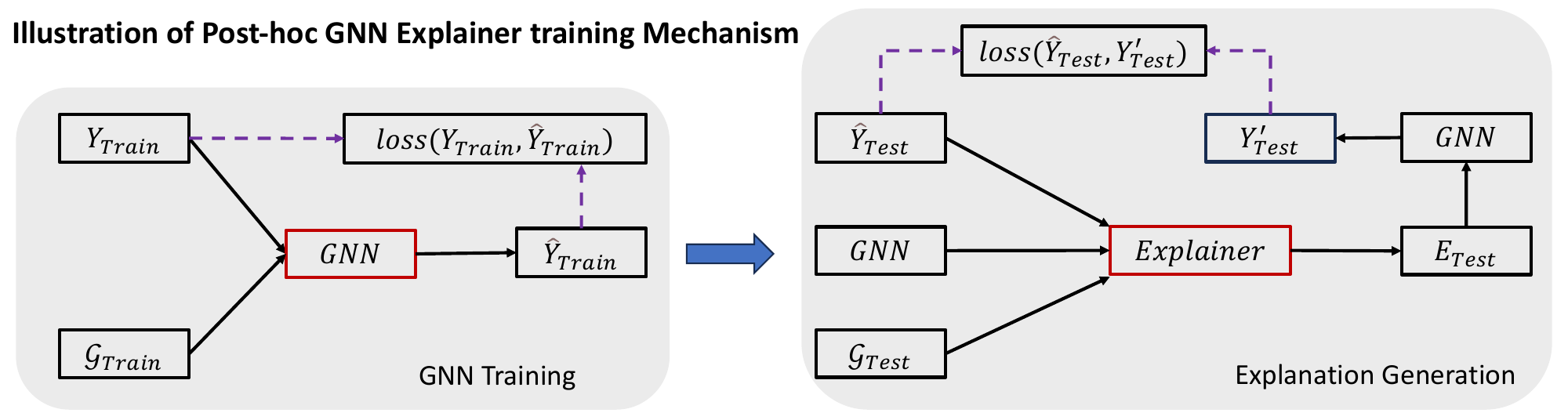}
\caption{
An overview of the workflow of selected post-hoc GNN explainers. 
}
\label{fig:post-hoc_gnnexplainer_mechanism}
\end{figure}

\begin{table}[!ht]
\caption{
    Evaluation of GNN models, GNNExplainer and PGExplainer on real-world datasets, with (w/) and without (w/o) data manipulation.
    For instance, $\lambda=$w/0.5 indicates 5\% training node labels are randomly manipulated. 
    Model performance is evaluated on classification accuracy (\ACC) and explanation quality is evaluated on fidelity+ (\FIDP) and fidelity- (\FIDN) on test dataset. 
}
\label{table:micro_evaluation_gnn_explainers}
\setlength{\tabcolsep}{3pt} 
\centering
\resizebox{1.\linewidth}{!}{
\begin{tabular}{l|c|cc|cc|cc|cc|cc|cc}
\toprule 
\textbf{Dataset} & $\lambda$ & \multicolumn{12}{c}{\textbf{GNNExplainer}} \\
\midrule
& & \multicolumn{6}{c|}{\GCN} & \multicolumn{6}{c}{\GIN} \\
& & \multicolumn{2}{c|}{\ACC} & \multicolumn{2}{c|}{\FIDP} & \multicolumn{2}{c|}{\FIDN} & \multicolumn{2}{c|}{\ACC} & \multicolumn{2}{c|}{\FIDP} & \multicolumn{2}{c|}{\FIDN} \\
\midrule
\midrule
\multirow{5}{*}{\MUTAG} 
& w/o 
& 0.818 & - & 0.375 & - & 0.425 & -
& 0.821 & - & 0.365 & - & 0.367 & - \\
& w/.5 
& 0.798 & -2.4\% & 0.337 & -10.1\% & 0.334 & -21.4\%
& 0.798 & -2.8\% & 0.344 & -5.8\% & 0.336 & -8.4\% \\
& w/.10 
& 0.790 & -3.4\% & 0.212 & -43.5\% & 0.215 & -49.4\%
& 0.794 & -3.3\% & 0.285 & -21.9\% & 0.276 & -24.8\% \\
& w/.15 
& 0.787 & -3.8\% & 0.166 & -55.7\% & 0.171 & -59.8\%
& 0.737 & -10.2\% & 0.208 & -43.0\% & 0.204 & -44.4\% \\
& w/.20 
& 0.778 & -4.9\% & 0.151 & -59.7\% & 0.156 & -63.3\%
& 0.786 & -4.3\% & 0.195 & -46.6\% & 0.192 & -47.7\% \\


\midrule
\multirow{5}{*}{\BAmotifs} 
& w/o 
& 0.980 & - & 0.496 & - & 0.496 & - 
& 0.980 & - & 0.486 & - & 0.486 & - \\
& w/.5 
& 0.990 & 1.0\% & 0.423 & -14.7\% & 0.397 & -20.0\%
& 0.985 & 0.5\% & 0.440 & -9.5\% & 0.439 & -9.7\% \\
& w/.10 
& 0.990 & 1.0\% & 0.415 & -16.3\% & 0.389 & -21.6\%
& 0.995 & 1.5\% & 0.364 & -25.1\% & 0.363 & -25.3\% \\
& w/.15 
& 0.970 & -1.0\% & 0.345 & -30.4\% & 0.342 & -31.0\%
& 0.965 & -1.5\% & 0.274 & -43.6\% & 0.274 & -43.6\% \\
& w/.20 
& 0.950 & -3.1\% & 0.248 & -50.0\% & 0.250 & -49.6\%
& 0.970 & -1.0\% & 0.231 & -52.5\% & 0.230 & -52.7\% \\

\midrule
\multirow{5}{*}{\texttt{BAMult.S.}} 
& w/o 
& 0.970 & - & 0.479 & - & 0.479 & - 
& 0.990 & - & 0.488 & - & 0.488 & - \\
& w/.5 
& 0.980 & 1.0\% & 0.392 & -18.2\% & 0.437 & -8.8\%
& 0.985 & -0.5\% & 0.530 & 8.6\% & 0.429 & -12.1\% \\
& w/.10 
& 0.980 & 1.0\% & 0.335 & -30.1\% & 0.313 & -34.7\%
& 0.980 & -1.0\% & 0.365 & -25.2\% & 0.363 & -25.6\% \\
& w/.15 
& 0.975 & 0.5\% & 0.294 & -38.6\% & 0.294 & -38.6\%
& 0.985 & -0.5\% & 0.334 & -31.6\% & 0.333 & -31.8\% \\
& w/.20 
& 0.990 & 2.1\% & 0.294 & -38.6\% & 0.298 & -37.8\%
& 0.960 & -3.0\% & 0.218 & -55.3\% & 0.228 & -53.3\% \\

\midrule
\multirow{5}{*}{\texttt{G.-Twitter}} 
& w/o 
& 0.635 & - & 0.605 & - & 0.594 & - 
& 0.665 & - & 0.286 & - & 0.262 & - \\
& w/.5 
& 0.612 & -3.6\% & 0.523 & -13.6\% & 0.515 & -13.3\%
& 0.651 & -2.1\% & 0.247 & -13.6\% & 0.229 & -12.6\% \\
& w/.10 
& 0.638 & 0.5\% & 0.244 & -59.7\% & 0.244 & -58.9\%
& 0.639 & -3.9\% & 0.261 & -8.7\% & 0.246 & -6.1\% \\
& w/.15 
& 0.623 & -1.9\% & 0.192 & -68.3\% & 0.189 & -68.2\%
& 0.620 & -6.9\% & 0.202 & -29.4\% & 0.192 & -26.7\% \\
& w/.20 
& 0.635 & 0.0\% & 0.186 & -69.3\% & 0.187 & -68.5\%
& 0.619 & -6.9\% & 0.194 & -32.2\% & 0.189 & -27.9\% \\

\midrule
\textbf{Dataset} & $\lambda$ & \multicolumn{12}{c}{\textbf{PGExplainer}} \\
\midrule
\midrule

\multirow{5}{*}{\MUTAG} 
& w/o 
& 0.809 & - & 0.307 & - & 0.431 & - 
& 0.823 & - & 0.330 & - & 0.482 & - \\
& w/.5 
& 0.807 & -0.2\% & 0.294 & -4.2\% & 0.366 & -15.1\%
& 0.780 & -5.2\% & 0.281 & -14.8\% & 0.308 & -36.1\% \\
& w/.10 
& 0.789 & -2.5\% & 0.249 & -18.9\% & 0.322 & -25.3\%
& 0.786 & -4.5\% & 0.261 & -20.9\% & 0.262 & -45.6\% \\
& w/.15 
& 0.787 & -2.7\% & 0.236 & -23.1\% & 0.316 & -26.7\%
& 0.767 & -6.8\% & 0.200 & -39.4\% & 0.193 & -60.0\% \\
& w/.20 
& 0.767 & -5.2\% & 0.206 & -32.9\% & 0.294 & -31.8\%
& 0.759 & -7.8\% & 0.144 & -56.4\% & 0.132 & -72.6\% \\


\midrule
\multirow{5}{*}{\BAmotifs} 
& w/o 
& 0.995 & - & 0.489 & - & 0.489 & - 
& 1.000 & - & 0.316 & - & 0.481 & - \\
& w/.5 
& 0.973 & -2.2\% & 0.380 & -22.3\% & 0.420 & -14.1\%
& 0.975 & -2.5\% & 0.310 & -1.9\% & 0.446 & -7.3\% \\
& w/.10 
& 0.971 & -2.4\% & 0.320 & -34.6\% & 0.331 & -32.3\%
& 0.995 & -0.5\% & 0.282 & -10.8\% & 0.378 & -21.4\% \\
& w/.15 
& 0.970 & -2.5\% & 0.284 & -41.9\% & 0.285 & -41.7\%
& 0.985 & -1.5\% & 0.004 & -98.7\% & 0.343 & -28.7\% \\
& w/.20 
& 0.950 & -4.5\% & 0.267 & -45.4\% & 0.282 & -42.3\%
& 0.980 & -2.0\% & 0.023 & -92.7\% & 0.240 & -50.1\% \\

\midrule
\multirow{5}{*}{\texttt{BAMult.S.}} 
& w/o 
& 0.980 & - & 0.482 & - & 0.482 & - 
& 0.985 & - & 0.281 & - & 0.485 & - \\
& w/.5 
& 0.965 & -1.5\% & 0.258 & -46.5\% & 0.360 & -25.3\%
& 0.980 & -0.5\% & 0.257 & -8.5\% & 0.438 & -9.7\% \\
& w/.10 
& 0.970 & -1.0\% & 0.246 & -49.0\% & 0.253 & -47.5\%
& 0.985 & 0.0\% & 0.236 & -16.0\% & 0.288 & -40.6\% \\
& w/.15 
& 0.955 & -2.6\% & 0.228 & -52.7\% & 0.229 & -52.5\%
& 0.985 & 0.0\% & 0.102 & -63.7\% & 0.326 & -32.8\% \\
& w/.20 
& 0.958 & -2.2\% & 0.218 & -54.8\% & 0.218 & -54.8\%
& 0.985 & 0.0\% & 0.051 & -81.9\% & 0.271 & -44.1\% \\

\midrule
\multirow{5}{*}{\texttt{G.-Twitter}} 
& w/o 
& 0.633 & - & 0.209 & - & 0.333 & - 
& 0.656 & - & 0.275 & - & 0.255 & - \\
& w/.5 
& 0.633 & 0.0\% & 0.168 & -19.6\% & 0.213 & -36.0\%
& 0.638 & -2.7\% & 0.192 & -30.2\% & 0.175 & -31.4\% \\
& w/.10 
& 0.638 & 0.8\% & 0.149 & -28.7\% & 0.186 & -44.1\%
& 0.629 & -4.1\% & 0.191 & -30.6\% & 0.174 & -31.8\% \\
& w/.15 
& 0.619 & -2.2\% & 0.125 & -40.2\% & 0.178 & -46.6\%
& 0.632 & -3.7\% & 0.191 & -30.6\% & 0.188 & -26.3\% \\
& w/.20 
& 0.594 & -6.2\% & 0.124 & -40.7\% & 0.161 & -51.7\%
& 0.596 & -9.2\% & 0.186 & -32.4\% & 0.176 & -31.0\% \\

\bottomrule
\end{tabular}
}
\end{table}




\end{document}